# Self-prompting and cross-model consensus enable reproducible data extraction from scientific literature with large language models

Valentin Romanov[1], Monique Bax[2,3], Steven Niederer[1,4]

1. National Heart & Lung Institute, Faculty of Medicine, Imperial College London, United Kingdom
2. Cambridge Stem Cell Institute, Jeffrey Cheah Biomedical Centre, Cambridge Biomedical Campus, University of Cambridge, Cambridge, United Kingdom
3. Department of Medicine, University of Cambridge, Cambridge, United Kingdom
4. Division of Cardiovascular Medicine, Stanford Cardiovascular Institute, Stanford University, CA, USA

**Abstract**

Accurately extracting nuanced, contextualized data from research articles is laborious and time intensive. Here, we investigate the performance of frontier, browser-based large language models (LLMs) to extract highly contextualized information. We demonstrate four escalating workflows, 1) given an expert curated prompt and research articles, most frontier LLMs perform well at data extraction, however can struggle with interpreting scientific context and nuance, 2) given simple instructions, LLMs can author their own prompts which were almost as effective as expert-written prompts, 3) autonomous discovery of research literature was difficult, agents either missed or hallucinated references, and 4) LLMs can create new datasets from published guidelines that closely match human-expert judges, but still require a human-in-the-loop. Together, these findings define an auditable division of labour in which experts specify the evidence standard, models cross-check repeated extractions and researchers resolve disputed cases, providing a practical route to scaling scientific data curation without relinquishing expert oversight.

## Introduction

Data extraction from unstructured research articles is time and labour intensive. A researcher may spend months combing through literature to find provenance of published parameters, and even then, extraction errors, miscalculations and misunderstanding of research nuance can pollute results[1]. While Large Language Models (LLMs) can significantly improve this process, with the potential to increase parameter extraction accuracy and efficiency[2], universal principles for how to do this well are still being developed[3,4].

As model capabilities develop, it becomes necessary to consider where LLMs are utilized. The difference between application programming interface (API) calls and a browser-session is significant. A browser user interacts not with an isolated LLM (with the ability to define a system prompt, temperature etc), but with a deployed product that may combine hidden instructions, document processing, retrieval, contextual memory, moderation and provider-controlled tools. LLM capabilities and performance also change across interface updates, and with compute allocation. Direct comparisons show that web user interface (WebUI) and API access can produce different behaviour even when they ostensibly expose the same model family[5,6]. Results obtained through APIs accordingly cannot automatically be generalized to consumer-facing web products.

Not only is there a divergence at the interface level, LLM performance depends on both the prompts they receive and the context they are given. Liu et al. (2025) found that long, detailed prompts improved performance on financial tasks,[7] whereas Levy et al. (2024) found that longer prompts reduced performance[8]. Liu et al. (2023) also showed that prompt length and the position of key information affected results[9], while Du et al. (2025) found that length mattered more than prompt structure[10]. Automated methods such as APE AI and OPRO can produce high quality prompts, but generally require labelled examples or other strategies for judging performance[11–16]. Even with well-grounded prompts and managed context[17], LLMs can still fail at both finding relevant research articles and to accurately extract data from them. Clark et al. (2025) found that LLM search workflows missed 68–96% of relevant studies and extracted 4–31% of items incorrectly[18]. Although new benchmarks and Deep Research approaches test or combine searching, citation checking and data extraction[19–23] evidence of their performance on real research tasks remains limited[24–26].

Additionally, automated literature extraction must be accurate, reproducible, and be able to be evaluated against a clear reference standard. This is challenging because papers often omit key methods, for example, it was found that preclinical studies reported fewer than half of the recommended Animal Research: Reporting of In Vivo Experiments (ARRIVE) items, on average[27]. In addition to human omissions, LLM output can also vary across runs and models. One way to improve LLM output is to obtain a consensus[28], where disagreements between LLMs may actually help flag difficult cases for human-review[29–31]. Although LLMs have been used to assess clinical-trial reporting at scale[32] and to support research-reporting and reproducibility checklists in machine learning and natural-language processing[33,34], it remains unclear how accurately and consistently LLMs can apply detailed reporting guidelines in other fields due to the lack of examples for how to develop prompts, how to work with different frontier models and how to successfully incorporate a human-in-the-loop.

Here, we evaluate four strategies that progressively increase model autonomy in extracting nuanced information from unstructured research articles. By utilizing the browser interface, through which most researchers and people access LLMs, we demonstrate a number of tactics and techniques for significantly improving LLM output across challenging data extraction tasks. We show that (1) with an expert-designed prompt and fixed research corpus, frontier LLMs approach human-level extraction performance; (2) LLMs can draft near-expert prompts but still require a human-in-the-loop to provide guidance and nuance; (3) deep-research agents vary markedly, with most diverging from the ground truth dataset we task them to replicate; and (4) a human-in-the-loop workflow, which uses the techniques demonstrated here, can be used to direct LLMs to generating a new dataset in an unrelated scientific field. Together, our findings provide a practical framework for deciding what can be delegated to LLMs, and where expert judgement (human-in-the-loop) remains essential.

## Results & Discussion

### Data extraction and curation using LLMs

As an initial case study, we used the review by Niederer et al. (2006), which collated measurements from multiple experimental studies, that measured the affinity of $Ca^{2+}$ binding to cardiac troponin C, that were used to inform a biophysical model of cardiac contraction[35] (**Fig. 1a**). Reconstructing the evidence base in this review is particularly challenging because many of the sources, which date back to the 1980s, often have incomplete or brief Methods sections, use nuanced era-specific science terms, and data that exists across tables and figures. For the LLM to identify the correct measurement, it must understand the experimental context of that measurement, including the pH, temperature, [$Mg^{2+}$] concentration, protein form and how it was measured.

Adding to this challenge is how to effectively use an LLM to extract this data. Performance depends on the level of detail and instruction in the prompt, whether the prompt is appropriate to that model, how its accessed (local vs API vs browser), the model tier (free and paid models differ in ability and context limits), whether the model is appropriate for the task, and the fact that LLMs are non-deterministic and can return different answers to the same query (**Fig. 1b**).

Here, using seven frontier LLMs: GPT-5.5, Opus 4.7, Gemini 3.1 Pro, Qwen3.7, Kimi K2.6, GLM-5.1 and DeepSeek v4, we develop and test four strategies of increasing complexity, each delegating more of the task to the model. First, each model receives the same human-expert curated prompt, as in most studies, with retrieval scored against the gold standard as the median or best of five (**Fig. 1c**). Second, we test whether an LLM can design an expert-level prompt, starting from an informed-user prompt that conveys only the researcher's intent (**Fig. 1d**). Third, we remove prompt engineering entirely and task a Deep Research agent with the full workflow, that is, finding references as well as extracting values (**Fig. 1e**). Lastly, we create a new dataset. We utilized published guidelines and an informed-user prompt to generate an LLM-specific master prompt, which is then used to extract data from articles in an unrelated field. We compare this against annotations solicited from researchers, focusing on where and how human-experts deviate from LLM findings (**Fig. 1f**).

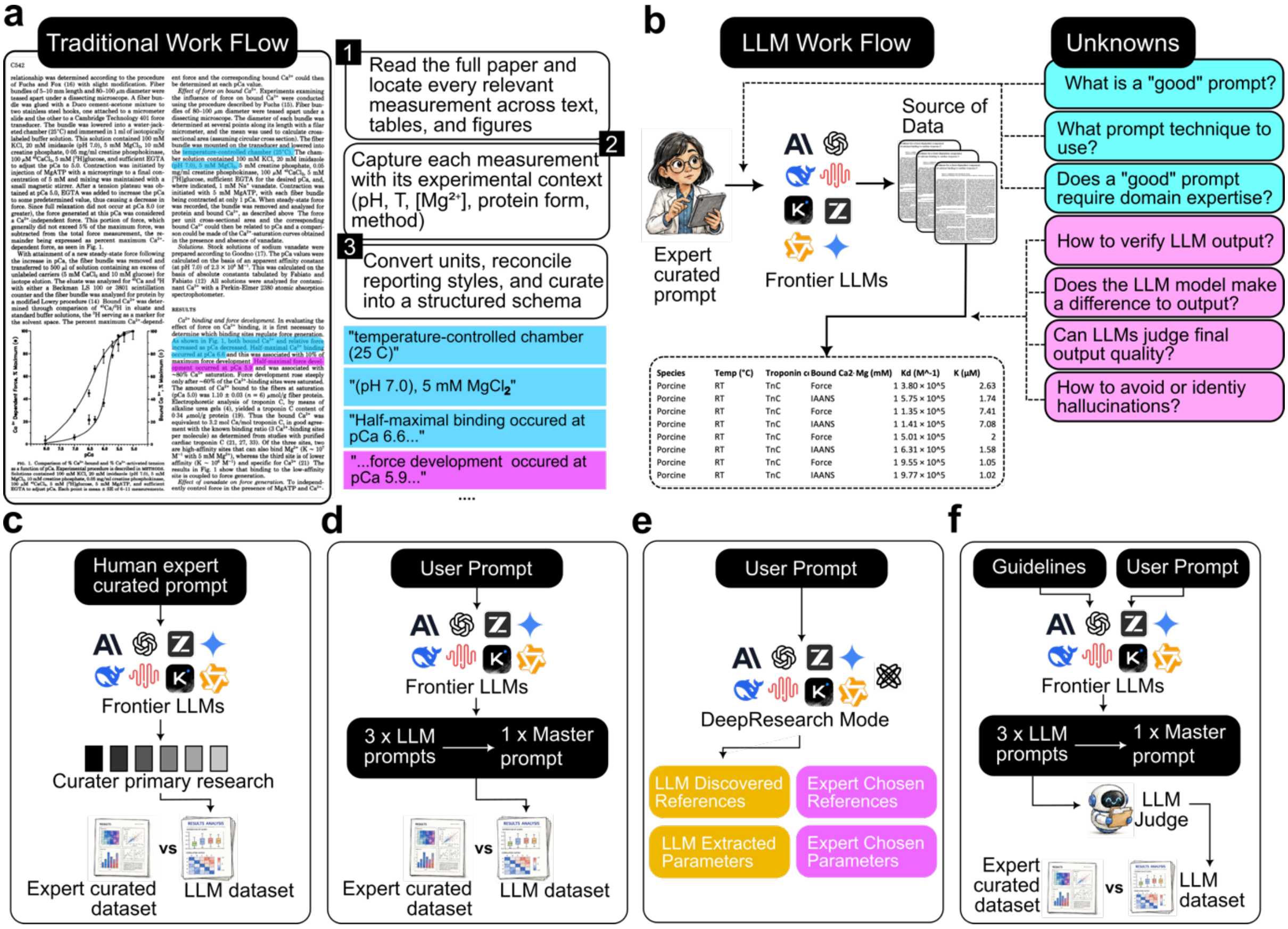

**Fig. 1: Framework and study design for LLM-assisted literature extraction. a,** Traditional manual data-extraction workflow. A researcher reads the full article, locates relevant measurements across the text, tables and figures, records the experimental context of each measurement, and reconciles units and reporting conventions to produce a structured dataset. **b,** General LLM-assisted workflow. Typically, articles and an expert-curated prompt are provided to the LLM to generate a structured dataset. However, many questions remain about how to do this well. **c,** First, we explore how a human-expert prompt performs with multiple frontier LLMs, and the resulting datasets are compared with an expert-curated reference dataset. **d,** Second, we investigate how frontier LLMs self-author prompts and aggregate them into a single master prompt, and how this prompt performs in data extraction relative to the expert-curated dataset. **e,** Next, we use autonomous deep-research agents to identify relevant primary literature and extract target parameters, comparing performance against the ground-truth dataset. **f,** Finally, we apply the concepts from c, d and e, to generate a new dataset in an unrelated scientific domain. Published guidelines and a user prompt are used to produce three model-authored prompts and an aggregated master prompt, after which the LLM-generated dataset is compared with human-judges.

## Expert curated prompt vs frontier-models

We asked whether frontier LLMs could follow a detailed, expert-written prompt to extract scientific data reliably. The prompt (**Supporting Prompt 1**) specified what information to find, how to organise it, and how to include the researcher-in-the-loop. We gave this prompt to each model alongside each of the 18 research articles, repeating every extraction five times (**Fig. 2a**). For every extraction, we checked whether the model correctly reported seven pieces of information: species, temperature, troponin preparation, measurement method, magnesium concentration and the two affinity values, Kd and K. GPT-5.5 performed best, correctly reporting 95.7% of the expected values and experimental conditions, followed by Gemini 3.1 Pro and Opus 4.7 at 91.4% and 90.6%, respectively (**Fig. 2b**). The three top models (GPT, Opus and Gemini) had an average accuracy of 92.6 ± 2.7%, compared with 86.4 ± 3.1% for the other four models. GLM-5.1 was the best performing open-source model. We note that model capabilities are improving rapidly, for example, DeepSeek V4 scored 17.6 percentage points higher than DeepSeek V3.2, partially due to better instruction-following (**Supporting Fig. S1**).

Significantly, in the process of extracting data and comparing it with the ground truth (**Table 1**), LLMs were able to identify several errors in the original dataset (**Supporting Table 3**). These errors included the wrong Kd to K conversion factor, incorrect temperature (25°C → RT) and incorrectly associated magnesium (Mg) values. These errors were easy to spot as almost all LLMs deviated from the ground truth. Using multiple LLMs to double check extraction quality, across multiple runs proved exceedingly effective and efficient.

LLMs performed well at extracting affinity data from research articles (**Fig. 2c**). The top models averaged 97.1 ± 2.6% while the rest of the models, 93.0 ± 4.1%, with GPT-5.5 reaching 99.5%. Models were therefore generally capable of finding or calculating an affinity value, but less reliable at determining exactly which biological preparation and experimental conditions that value belonged to. Models also frequently returned additional results that could not be matched to the reference dataset (**Fig. 2d**). Kimi K2.6 extracted an average of 3.6 additional entries per expected result yet recovered only 89.2% of them; Opus 4.7 produced the fewest (1.9) while recovering 95.4%, and GPT-5.5 recovered almost everything (99.5%) while still producing 2.7. The extra output was data directly extracted from the article but did not answer the original question. The model hedged on extracting more data than surgically finding the requested information. More output therefore did not necessarily mean better accuracy, it often meant more material that a researcher would have to check manually.

Consistency was an issue even among frontier models (**Fig. 2e**). Across the seven models, only three quarters (76.4%) of the extracted parameters were correct in all five runs, 4.4% wrong in all five and 19.2% changed between correct and incorrect. The three leading models gave the same correct answer in all five runs for 87.7% of parameters, compared with 67.9% for the other four (**Table 1**). A single run can misrepresent the capabilities of a model. Kimi K2.6 achieved 95.2% in one run and only 78.4–80.2% in the other four. We therefore recommend running each prompt a minimum of five times and reporting the 5 run majority.

Interestingly, the LLMs were able to find and extract the numerical parameters (K and Kd) correctly in 99.0% and 99.8% of cases, respectively, but identified the correct troponin preparation, which required deeper reasoning, in only 80.7% of cases (**Fig. 2f**). Extraction difficulty was not spread evenly across the 18 articles (**Fig. 2g**). The lowest average scores occurred for Hofmann and Fuchs (1987; ref33; 62.4 ± 30.1%), Pan and Solaro (1987; ref28; 66.1 ± 29.8%) and Parsons et al. (1997; ref29; 75.5 ± 8.0%; mean ± SD across models). On Hofmann and Fuchs, Opus 4.7 scored 0% and Kimi K2.6 94.3%; on Pan and Solaro, Gemini 3.1 Pro scored 17.1% and GPT-5.5 100%. Parsons et al. was more consistently difficult, ranging from 60.0% to 82.9%. The three most difficult papers (ref28**:** Pan and Solaro (1987), ref29**:** Parsons et al. (1997) and ref33**:** Hofmann and Fuchs (1987))[36–38] required models to reason over measurements made in skinned fibres or native myofilaments, separate those taken at different magnesium concentrations, and work out from the methods how each was performed and sometimes calculate an affinity that was not reported directly in a table. While extraction quality strongly depends on model "intelligence" and a well-curated prompt, it also depends on the article, for example, in the hardest papers, the same affinity was reported by two different methods, the reference value appeared in the body of the text while the tables listed other, sometimes contradictory constants and two of the papers downloaded were only available as images, in which optical character recognition (OCR) corrupted the exponents (**Fig. 2g**). Another interesting point of failure was that articles written in the 1980s used different nomenclature, which is a problem for an LLM that was specifically trained to faithfully follow instructions, like Opus 4.7, if the prompt did not account for this (list of research articles used in Fig. 2g – Supporting info).

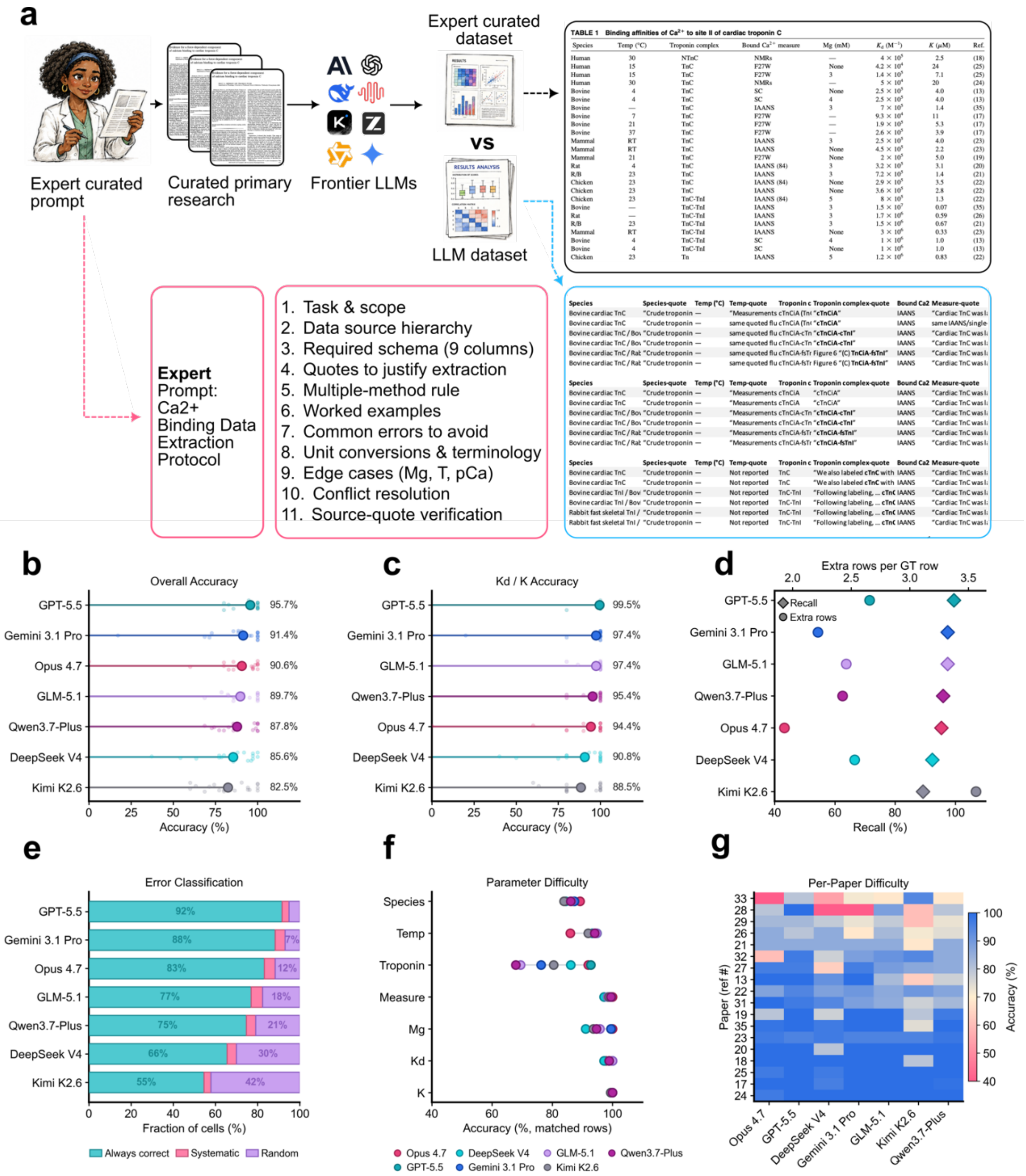

**Fig. 2: Frontier LLM performance on Ca2+ binding data extraction from primary research articles.** **a**, Overview of the benchmark workflow. An expert-curated Ca2+ binding data extraction prompt was provided to each model alongside each article, and model outputs were cleaned, aligned to a fixed schema, and scored against a corrected ground-truth dataset. **b**, Overall extraction accuracy across all scored fields. Large points show mean model accuracy across five independent runs; small points show per-paper performance. **c**, Accuracy restricted to the quantitative affinity fields, Kd and K. **d**, Extraction efficiency, measured as ground-truth row recall and the number of extra rows generated per ground-truth row. **e**, Error classification across repeated runs. “Always correct” indicates fields extracted correctly in all five runs, “systematic” indicates fields missed in all five runs, and “random” indicates inconsistent extraction across runs. **f**, Parameter-level difficulty across models, showing that contextual fields such as troponin context were more difficult than numerical affinity fields. **g**,

Per-paper difficulty heatmap, showing mean accuracy for each article-model pair across five runs. Pink indicates lower accuracy and blue indicates higher accuracy. Panels b-g show 7 frontier models evaluated on 18 primary research articles, with 5 independent runs per article.

| | Extraction turn | | | | | Summary | | |
|---|---|---|---|---|---|---|---|---|
| **Model** | **1** | **2** | **3** | **4** | **5** | **Best of 5** | **5-run Majority** | **Spread (pp)** |
| **GPT-5.5** | 96.7 | 96.3 | 94.1 | 95.6 | 95.6 | 96.7 | 96.7 | 2.6 |
| **GLM-5.1** | 92.3 | 90.1 | 92.7 | 80.2 | 93.4 | 93.4 | 93.0 | 13.2 |
| **Opus 4.7** | 90.5 | 90.1 | 89.4 | 90.5 | 92.7 | 92.7 | 92.3 | 3.3 |
| Gemini 3.1 Pro | 93.8 | 90.8 | 91.6 | 89.0 | 91.9 | 93.8 | 91.2 | 4.8 |
| Qwen3.7-Plus | 88.3 | 79.9 | 91.9 | 91.2 | 87.9 | 91.9 | 89.4 | 12.1 |
| Kimi K2.6 | 79.5 | 78.4 | 95.2 | 80.2 | 79.1 | 95.2 | 89.0 | 16.8 |
| DeepSeek V4 | 88.6 | 87.9 | 75.5 | 82.8 | 93.0 | 93.0 | 87.9 | 17.6 |

**Table 1. Accuracy for selected models across five independent extraction runs.** Each run is scored against the curated ground truth. Best of 5 is the highest accuracy achieved in any single run. 5-Run Majority reports the fraction of expected ground-truth fields for which the model extracted the correct value in at least 3 of 5 independent runs. Spread (pp) is the difference between the model's best and worst single-run accuracy, in percentage points, showing run-to-run variability. Colour scale: blue, highest accuracy; red, lowest accuracy.

**LLM self-authored prompts and the master prompt**

We next asked each model to design its own data-extraction prompt. The instructions required two preparatory searches: one for prompt-engineering practices appropriate to the model, and one for field-specific terminology in the relevant literature. We also specified the data, how to format the output table and how to justify anything the LLM found (**Supporting Prompt 2**). Each model generated three candidate prompts and then combined them into a single master prompt (**Supporting Prompt 3**), which was applied in five independent runs to eight research articles spanning the "easy", "medium" and "hard" categories defined in Figure 2g (**Fig. 3a**). We assessed two things, how closely each master prompt reproduced the expert-designed gold-standard prompt (see **Figure 2**), and how accurately the model extracted with it.

Each prompt was scored against a 38-component rubric covering task definition, source prioritization, required columns and quotations, calculations, special cases, worked examples and conflict resolution instructions, found in the expert-curated prompt. Aggregating prompts into a single master prompt improved the rubric score for every model. GPT-5.5 achieved the highest score, increasing from an average of 82.9% across its three individual prompts to 88.2% for its master prompt; Gemini 3.1 Pro and Kimi K2.6 gained the most, 11.4 and 10.5 percentage points, respectively (**Fig. 3b**). We then applied each master prompt to the eight articles. GPT-5.5 achieved the highest overall accuracy at 90.3% and Opus 4.8 the next at 85.5%, with the other five models between 57.8% and 74.5% (Fig. 3c). Combined association- and dissociation-constant accuracy exceeded overall accuracy for every model, reaching 98.1% for GPT-5.5 (**Fig. 3d**). GPT-5.5 was also the most consistent, returning accuracies of 90.1–90.7% across its five runs (SD, 0.3 pp) (**Supporting Table1**).

How closely the LLM prompt matched the expert-prompt did not strongly predict extraction accuracy. The rankings were only weakly and non-significantly associated across the seven models (Spearman $\rho$=0.36, $P$=0.43). DeepSeek V4 produced the second-highest-scoring master prompt but ranked fifth for extraction accuracy, whereas Opus 4.8 produced the fourth-highest but ranked second (**Supporting Fig. S2**). Compared with the expert prompt, the model-generated master prompts reduced overall extraction accuracy for every model (**Fig. 3e**). GPT-5.5 remained the most accurate model, falling 5.3 pp from 95.6% with the human-expert prompt to 90.3% with the LLM master prompt; Opus 4.8 fell 4.8 pp and Gemini 3.1 Pro 32.5 pp. During testing, Gemini 3.1 Pro did not perform the requested web searches despite repeated prompting, which is the reason why its master prompt had lower overall coverage against the human-prompt and why it inevitably performed worse.

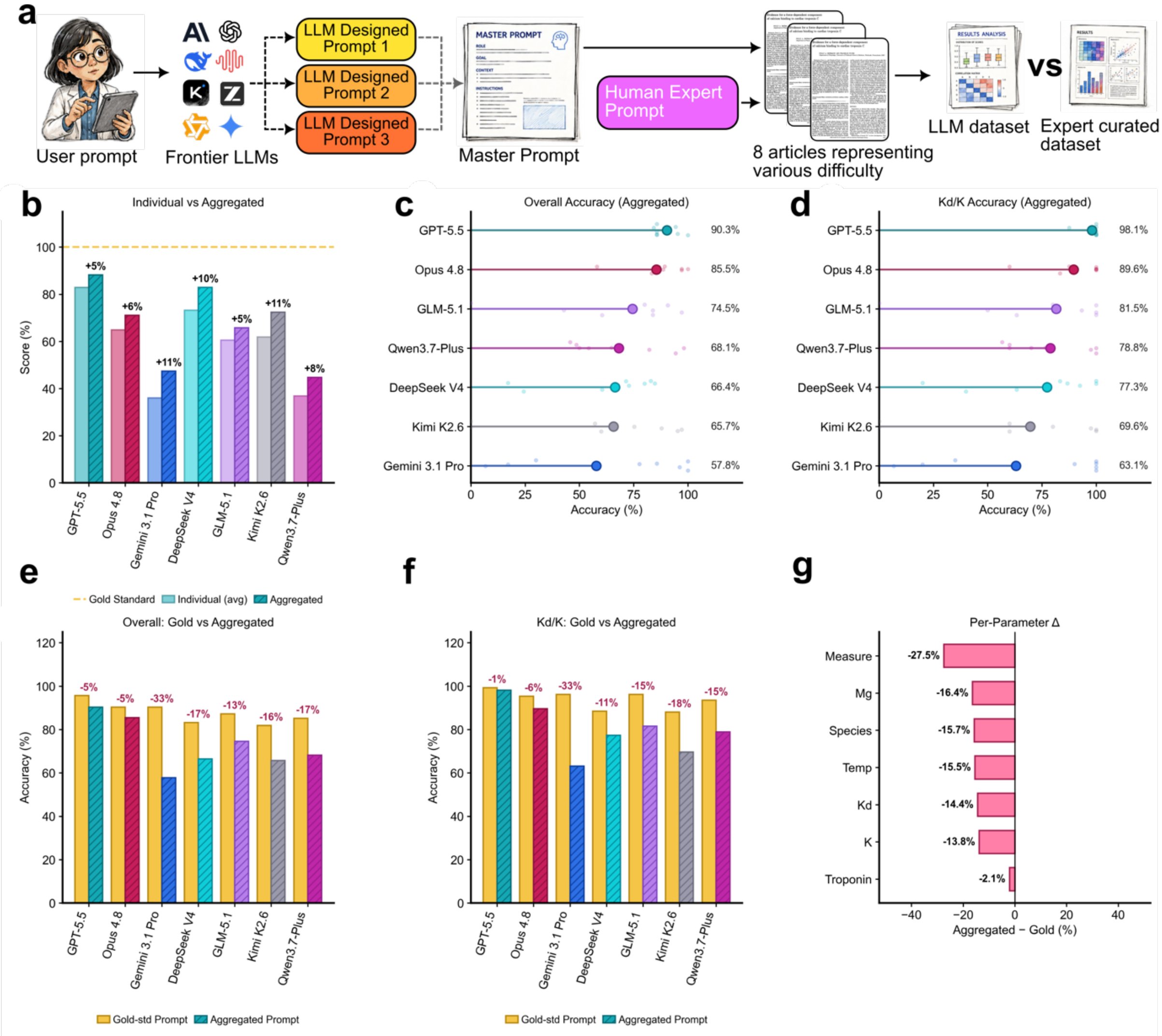

**Figure 3. LLM-authored prompts vs human-expert curated prompt. a**, Overview of the prompt-engineering workflow. Each model was instructed to search the internet for model-specific prompt-engineering practices and the primary literature for domain-specific terminology, then to generate three candidate extraction prompts and aggregate them into a single master prompt. Each master prompt, alongside the human expert gold-standard prompt from Figure 2, was applied to the research articles. **b**, Prompt quality scored against the gold-standard prompt. For each model, bars show the mean of its three individual prompts (solid) and its aggregated master prompt (hatched); labels give the aggregation gain, and the dashed line marks the gold-standard prompt (100%). **c**, Overall extraction accuracy using the aggregated master prompt. Large points show mean model accuracy across five independent runs; small points show per-paper performance. **d**, Accuracy restricted to the quantitative affinity fields, Kd and K, under the aggregated master prompt. **e**, Overall extraction accuracy under the gold-standard prompt (gold) versus the aggregated master prompt (hatched). **f**, The same comparison restricted to the Kd and K fields. **g**, Parameter-level change in accuracy from the gold-standard to the aggregated master prompt, averaged across models; negative values indicate fields extracted less accurately under the self-authored prompt. Each model was evaluated against eight primary research articles, with five independent runs per article.

Affinity extraction was more robust to prompting. GPT-5.5's combined association- and dissociation-constant accuracy decreased by only 1.1 pp, from 99.2% with the human-expert prompt to 98.1% with the LLM master prompt (**Fig. 3f**). We therefore decomposed the difference between the expert and master prompts by extraction field (**Fig. 3g**). The measurement-method field showed the largest mean reduction in accuracy, 27.5 pp across models, while magnesium concentration, species, temperature and the two affinity constants fell by 13.8 to 16.4 pp. Troponin-complex classification was comparatively preserved, decreasing by only 2.1 pp. These results are consistent with the expert prompt providing more explicit parameter-specific instructions, particularly for interpreting experimental measurement methods.

**Deep Research agents**

We asked whether each agent could locate the primary literature underlying the gold standard. We ran each model five times in its deep-research configuration using the same LLM-authored prompt (**Supporting Prompt 3 and 4**), allowing each system to search, browse and compile an independent report (**Fig. 4a**). Each run was instructed to identify up to 20 primary research articles published before 2005. Opus 4.8 returned the most citations, averaging 20.6 ± 1.4 per run, while Mistral 3.5, the only European model tested, returned the fewest, with 4.4 ± 1.6 citations per run (**Fig. 4b**).

We next asked how many of the 19 primary references underlying Niederer et al. 2006 Table 1 were recovered by each model. Opus 4.8 recovered the most, at a mean of 53.7% ± 14.7 percentage points. GPT-5.5 followed at 38.9% ± 2.6 pp and the dedicated literature-search agent Edison QA3 a t 29.5% ± 2.6 pp (**Fig. 4c**). Thus, even the leading Deep Research model missed approximately half of the expert-curated references. Article access may have contributed to this gap. Overall, 63% of the real papers cited were publicly accessible, whereas 37% were closed or paywalled. The open-access fraction was highest for Edison at 91%, followed by GLM-5.2 at 72% and Gemini 3.1 Pro at 67% (**Supporting Table 2**).

We then verified every unique citation and classified it as either accurate and verified, referring to a real paper but containing an error, or hallucinated (**Fig. 4d**). Qwen3.7-Plus and GLM-5.2 performed poorly on this measure: Qwen hallucinated 45% of its unique citations (27 of 60), while GLM-5.2 hallucinated 29% (20 of 69). The other five agents; GPT-5.5, Opus 4.8, Gemini 3.1 Pro, Mistral 3.5 and Edison QA3 produced no hallucinated citations. Only 2 of Qwen3.7-Plus's 60 unique citations were verified without error; 31 referred to papers but contained errors in the authors, year, journal or some combination of these, and 27 were completely hallucinated.

Report length was not a reliable indicator of extraction quality. Edison produced the longest reports, averaging 10,035 ± 2,466 words and reaching 14,441 words in one run, approximately twice the mean length of the Opus 4.8 and GPT-5.5 reports (**Fig. 4e**). Across the seven-model cohort, mean report length and mean field accuracy were only moderately correlated (r=0.51). While Opus 4.8 had the highest reference recall against the ground truth, it was not consistent, recovering 37–74% of the reference set across runs (SD 14.7 pp), while Gemini 3.1 Pro recovered 5–37% (SD 10.3 pp).

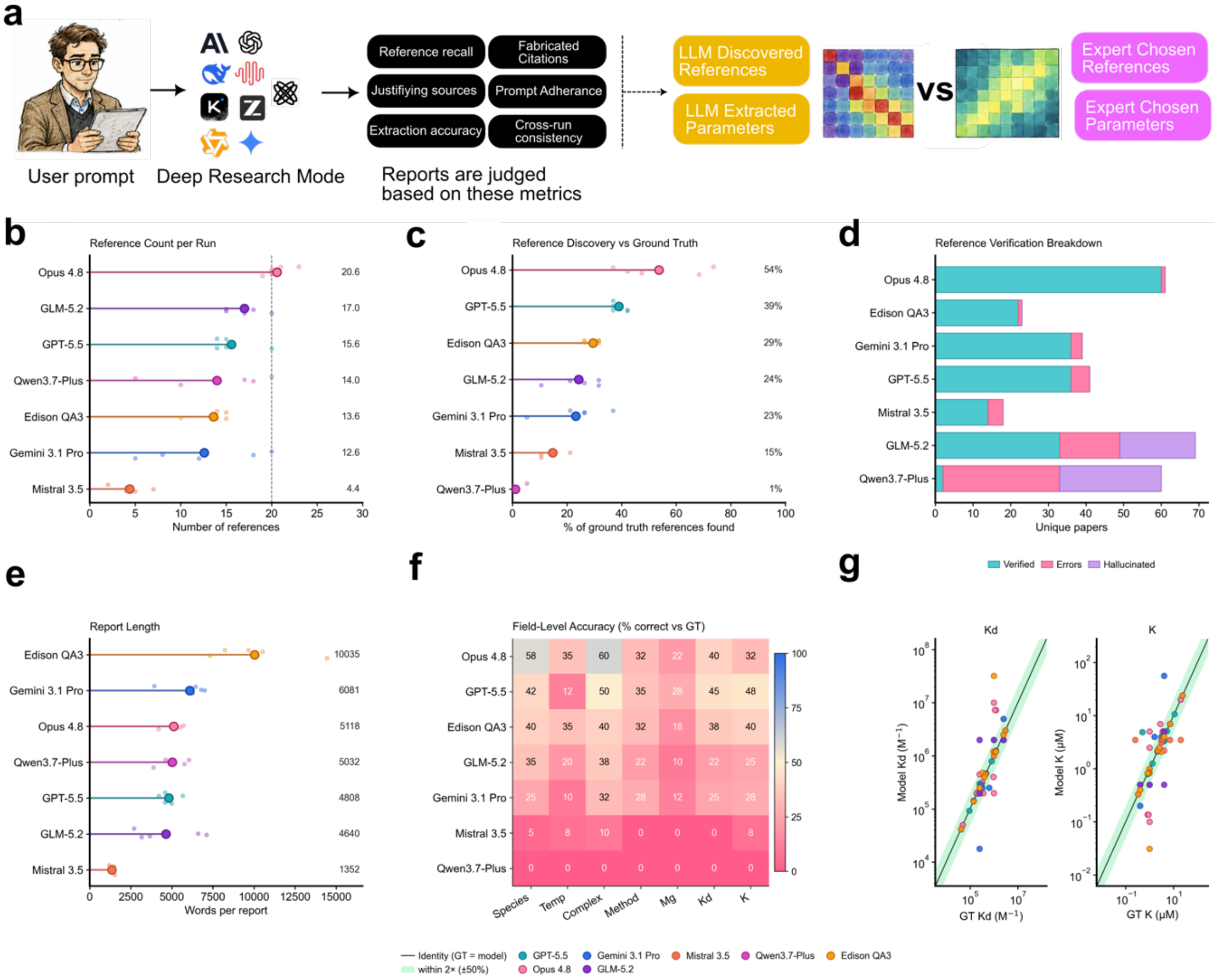

**Figure 4. Literature discovery and calcium-binding data extraction by Deep Research agents. a**, Overview of the evaluation workflow. Seven agents independently received the same LLM-authored prompt in five runs, with instructions to identify up to 20 primary research articles published before 2005 and extract cardiac troponin C calcium-binding data. Outputs were compared with the 19-reference, 40-condition ground truth derived from Table 1 of Niederer et al. (2006). **b**, Number of references reported per run. Small points denote individual runs, large points denote five-run means and the dashed line indicates the target of 20 references. **c**, Percentage of the 19 ground-truth references recovered in each run. Small and large points denote individual runs and five-run means, respectively. **d**, Verification status of unique papers cited across the five runs. Citations were classified as: verified without error, referring to a real paper but containing a citation error, or hallucinated. **e**, Report length measured in words. Small points denote individual runs and large points denote five-run means. **f**, Field-level extraction accuracy after pooling each model's five runs. Values show the percentage of scorable ground-truth conditions correctly recovered for species, temperature, troponin-complex state, experimental method, magnesium concentration, association

constant and dissociation constant. **g**, Agreement between model-reported and ground-truth constants K and Kd. The solid line indicates perfect agreement between the model and ground truth; points above or below it represent overestimation or underestimation, respectively. Both axes are logarithmic.

Each agent was also asked to extract the reported calcium-binding parameters from each deep research report. Averaged across the seven scored fields, accuracy against the ground truth reached 39.9% for Opus 4.8, 37.1% for GPT-5.5 and 34.6% for Edison QA3 (**Fig. 4f**). No model exceeded 60% on any single field, Opus 4.8 led on species, temperature and troponin-complex state while GPT-5.5 recovered both affinity constants more often, 45% against 40% for Kd and 48% against 32% for K. When a model did recover an experimental condition and reported a numerical affinity value, that value was usually close to the ground truth. Among these, 89% of association constants K and 84% of dissociation constants Kd fell within a factor of two (**Fig. 4g**).

**Creating new data sets using the techniques developed in prior sessions**

We combine these prior learnings (from Figures 2, 3 and 4) to demonstrate how to build new datasets, asking whether frontier LLMs can identify and document recommended reporting practices from research articles similarly to human scientific reviewers. We tasked seven models and human-reviewers with PhD-level scientific training to assess 10 articles[39–48] against 23 minimum and ideal reporting criteria (**Fig. 5a**). We based these criteria on recommendations from Sandoval et al who developed a set of guidelines defining the standard reporting practices for quantitative analysis of organoids[49]. For example, from the criteria “Morphology and Lineage” the researcher should, at minimum, sample three to five organoids from each differentiation batch; ideally, prior experiments and an appropriate power analysis should determine the sample size (**Fig. 5b**).

We began by examining the level to which LLMs agreed with human-reviewers. Interestingly, we found an immediate complication with using human-reviewers. Because reporting was not always a straightforward Yes/No decision, we observed variation in assigning a Yes or No decision when the decision could go either way, that is, a high level of uncertainty. Reviewers assigned the same classification in 79.1% of assessments but differed in 20.9%. LLMs achieved a mean agreement of 83.9% with Reviewer 1 and 78.2% with Reviewer 2, giving an overall mean of 81.1%. Individual model–reviewer agreement ranged from 75.2% to 87.0%. For some models, changing the reviewer used for comparison shifted the agreement estimate by as much as 7.8 pp (**Fig. 5c**).

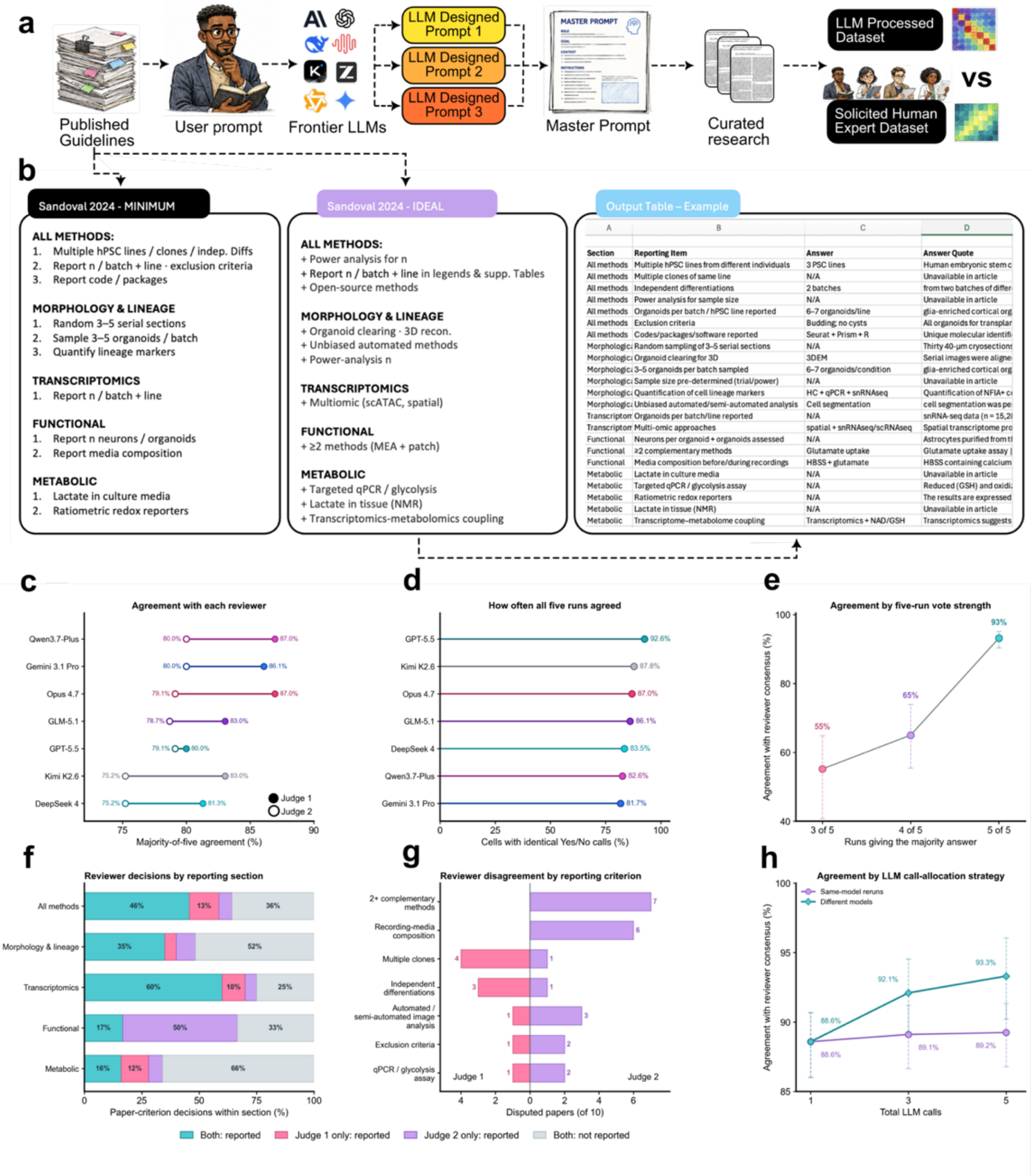

**Figure 5. Reproducibility and human agreement of LLM-extracted brain-organoid reporting criteria. a,** Overview of the self-prompting and extraction workflow. Each of seven frontier LLMs generated three candidate prompts from published organoid-reporting guidelines, aggregated them into a model-specific master prompt, and applied that prompt to 10 brain-organoid research articles in five independent runs. LLM-generated datasets were compared with assessments from PhD level researchers. **b,** The 23 reporting criteria derived from the minimum and ideal recommendations of Sandoval *et al.* (2024), grouped into all-methods, morphology and lineage, transcriptomic, functional and metabolic sections, with an example structured output containing the extracted answer and supporting quotation. **c,** Majority-of-five LLM agreement with each human reviewer. Filled and open points denote Judge 1 and Judge 2, respectively. **d,** Within-model reproducibility, measured as the percentage of extractions for which all five runs returned the same call, either Yes or No. **e,** Agreement

between the majority-of-five findings for all LLMs and alignment with reviewer findings. Results pool the seven models over all Yes calls, where reviewers agreed. Error bars: 95% CI. **f,** Human-reviewer decisions compared by each section found in the guidelines. Criteria is classified as reported by both reviewers, by Judge 1 only, by Judge 2 only or by neither reviewer. **g,** The seven reporting criteria with the greatest reviewer disagreement. **h,** Comparing agreement with reviewers based on 1 run for all models (teal), vs 5 runs per single model (purple). Error bars: 95% CI.

Agreement with a reviewer, however, did not tell us whether a model would produce the same answer again. We therefore ran each model five times using its own fixed prompt and measured how often all five Yes/No classifications were identical. GPT-5.5 was the most reproducible model at 92.6% ± 7.7 percentage points across papers, followed by Kimi K2.6 at 87.8% and Opus 4.7 at 87.0%. Gemini 3.1 Pro was the least reproducible at 81.7%, meaning that at least one of its five runs differed in 18.3% of assessments. Yet Gemini's majority-of-five classification still agreed with reviewer consensus in 91.8% of assessments, showing that reproducibility and reviewer agreement captured different aspects of model behaviour (**Fig. 5d**).

This contrast raised a second question: could disagreement among reruns serve as a warning that the resulting answer was less reliable? Pooling all models and assessments on which the reviewers agreed, a unanimous 5–0 model answer matched reviewer consensus in 93.2% of cases. Agreement fell to 65.0% when the runs split 4–1 and to 55.2% when they split 3–2 (**Fig. 5e**). As such, the strength of agreement among the five runs provided directionality about how likely the majority answer was to match the reviewers. As an interesting aside, one of the chosen articles was a Concept Paper[44]. Only Opus 4.7 informed us of this fact without being prompted. LLMs will typically proceed with the user's request, regardless of whether the underlaying context is inappropriate. It is up to the researcher to verify all sources.

We next returned to the human assessments to understand where reviewers disagreed. Disagreement was not distributed evenly across the reporting framework. Reviewer agreement was highest for "Morphology and Lineage" at 86.7% and "Transcriptomics" at 85.0%, but fell to 50.0% for "Functional criteria". Transcriptomics had the highest proportion that both reviewers classified as reported, at 60.0%, whereas Metabolic criteria had the highest proportion that both classified as not reported, at 66.0%. Functional disagreement was more polarizing, in 50.0% of Functional assessments, Reviewer 2 classified the criterion as reported whereas Reviewer 1 did not (**Fig. 5f**). In total, seven criteria accounted for 66.7% of all reviewer disagreements (**Fig. 5g**).

Finally, using a diverse set of frontier models allowed us to ask a practical question: if several LLM models are available, should researchers ask the same model to assess the paper repeatedly or ask different models and combine their answers? Asking the same model to repeat the assessment produced little change in agreement with reviewer consensus, topping at 89.2% with five runs. By contrast, majority voting across different models increased agreement to 92.1% with three runs and 93.3% with five, gains of 3.0 and 4.1 pp. For this benchmark, additional runs produced higher agreement when distributed across different models than when repeatedly assigned to the same model (**Fig. 5h**).

## Conclusion

This study demonstrates that reliable LLM-assisted literature curation depends not only on model intelligence, but also on prompt engineering, well curated sources and human-in-the-loop validation. We show that the current frontier models recovered more than 90% of ground-truth values and experimental conditions when given a fixed article set and an expert prompt. Numerical affinity constants approached 99% accuracy. The principal challenge was establishing the nuance and context under which numerical values should and shouldn't be extracted. Repeated runs also showed that a single run can misrepresent model performance.

We further show that LLMs can generate effective master prompts from an informed-user brief. Although these prompts did not match the performance of the expert-written prompt, they could provide a strong starting point for iterative user-refinement. Greater autonomy exposed a more substantial constraint. When deep-research agents were required to locate and interpret the source literature, even the strongest system recovered only 53.7% of the expert-curated primary references on average, and two systems generated hallucinated citations.

A central finding is that model consensus is an important part of LLM-output quality control. Applying repeated and cross-model assessments to an unrelated organoid-reporting dataset revealed disagreement not only between models, but also between the human reviewers used to establish the reference standard. Where reviewers agreed, model consensus closely reproduced their assessments; where models disagreed, those cases could be directed to expert review. Convergence across models also helped expose errors in the original cardiac reference dataset, showing that consensus can audit both extracted data and the standards used to evaluate them.

Together, these findings establish a practical framework for reproducible LLM-assisted literature curation. Experts define the question, sources and evidence standard; models perform repeated extractions across systems; and human review focuses on contested, contextual or high-impact cases. This division of labour provides a transparent and auditable route to constructing high-quality scientific datasets while retaining human verification, feedback and scientific judgement.

## Methods

### Models

Figures 2, 3 and 5 were completed with the following models: GPT-5.5 (xhigh), Claude Opus 4.7 (max), Gemini 3.1 Pro (extended thinking), DeepSeek V4 (instant), GLM-5.1 (Deep think, Max), Kimi K2.6 (Instant + High) and Qwen3.7-Plus (thinking). Where possible, we explicitly turned off the LLMs ability to access internet. Its unclear which model is served when using Deep Research mode as such, we list the model-era the analysis was performed in:  GPT-5.5 (Deep research), Claude Opus 4.8 (Research), Gemini 3.1 Pro (Deep research), GLM-5.2 (Advanced Search), Mistral 3.5, Qwen3.7-Plus (Deep research) and Edison Literature (High).

### Prompts

Prompts used to extract data, create new LLM authored prompts, and create master prompts can be found in the Supporting Information document.

### Browser access

Over the course of generating these results, not only have the models change but so has their behaviour in the browser. Initially, Gemini 3.0 Pro was very capable at searching and generating its own prompts, whereas we found that Gemini 3.1 Pro consistently refused to search, typically hallucinating that it did so. We had initially benchmarked MiniMax 3 but gave up as this platform eventually scaled down compute to free accounts. Similarly, when GLM-5.2 was first released very little compute was assigned to the web clients, as such, we could not perform our analysis. While Opus 5 is available as of this writing, we are unable to use it without draining tokens allowance immediately. Of the models, Qwen was the only platform where we could run 5 separate conversations at once, all day, without hitting limits.

### Generating initial LLM dataset

All models were accessed via their respective web browsers. For each model, we opened 5 conversations. We pasted the expert-prompt into each, added the pdf (one at a time) and collected the output for further processing.

We used the primary studies summarised in Table 1 of Niederer et al. (2006). The table contained 19 full-text primary articles and 40 experimental-conditions, of which, 18 were accessible to us. As such, the final corpus was 18 primary articles and 39 conditions. The reference data described species, temperature, troponin preparation, $Ca^{2+}$-measurement method, magnesium concentration, association constant Kd ($M^{-1}$) and dissociation constant K (μM).

In addition to extracting all parameters of interest, each parameter came with a quote to allow us to quickly validate against the original article and a nuance column. As the primary interest of the original review were the numerical parameters, we anchored extraction based on K and Kd values. If the model failed to locate either, it was classified as missing and scored incorrect for every field. A model row that could not be assigned to an expected row was classified as extra. A majority-of-five result was positive when at least three of the five runs were positive.

**LLM self-authored prompts and master prompt**

Each model was first instructed to research provider-specific prompting guidance and scientific terminology relevant to cTnC affinity extraction. It generated three candidate prompts, each intended to extract a structured table from one uploaded article while preserving supporting quotations and any calculations. For figures 3 and 5, every model was given the ability to search the internet: Opus 4.7 (max - web search), GLM-5.1 (Deep think + Max + Advanced Search), DeepSeek v4 (DeepThink + Search), Kimi k2.6 (Instant + High + web search – auto) and Qwen3.7-Plus (Thinking + Web search). For each model we opened 3 new conversations, and generated 3 initial prompts, combined these into one document. A subsequent aggregation instruction asked the model to combine the three candidates into one master prompt without discarding useful detail. The resulting model-specific master prompt was then used for five extraction runs on each of eight primary articles, representing 26 expected experimental-condition rows.

LLM designed master prompts were compared to the single expert curated prompt, decomposed into 38-components covering task definition, source prioritisation, required columns and evidence, calculations, special cases, worked examples and conflict resolution. Each component was scored 0 (absent), 1 (partially present) or 2 (fully present), giving a maximum score of 76. Initial scoring of master prompts was performed by Opus 4.8 (xhigh) followed by human validation, verification and confirmation.

**Deep Research Literature-Discovery Benchmark**

Seven Deep Research agents independently received the same master prompt in five separate runs. The initial, and master prompts can be found in Supporting Prompt S4 and S5. Each report was compared against the prompt for compliance. Every reference found in the report was manually verified, specifically focusing on whether the reference was real, and whether it was open-access.

**Brain-organoid reporting benchmark**

We converted the minimum and ideal reporting recommendations of Sandoval et al. (2024) into 23 paper-level items grouped into five sections: methods applicable to all studies (7 items), morphology and lineage (6), transcriptomics (2), functional assessment (3) and metabolism (5). Seven models each generated three candidate prompts, aggregated them into a model-specific master prompt, and applied that fixed prompt to ten brain-organoid articles in five runs. The 10 papers were hand selected to contain most of the features asked for in the guidelines. Data from LLMs was captured in an excel document, retaining the raw answers, supporting quotes and a binary Yes/No to whether the LLM flagged the article as having covered the particular guideline.

Two PhD-level scientific reviewers independently assessed the same ten papers and 23 items, scoring each criteria as either Yes, the articles addressed the guideline, or No. For each, the majority-of-five was Yes when at least three runs were Yes.

**Use of generative AI**

Generative AI (GPT-5.6 Sol (xhigh) and Fable 5 (high) + Opus 5 (med)) were used to writePpython scripts to process data and to copy-edit this manuscript.

**Data Availability**

Repository contains:

- All prompts used across this work
- Raw and curated data for all figures
- Tutorial to try out the workflow developed here for Figures 2 and 3

Repository: https://github.com/ValentinRoma26/romanov-2026-llm-extraction

**Acknowledgements**

The project is funded by the UKRI Engineering and Physical Sciences Research Council (EPSRC) via grant (EP/Z531297/1)